\documentclass[runningheads]{llncs}
\usepackage{hyperref}
\usepackage{graphicx}
\usepackage{multirow}
\usepackage{cite}
\usepackage{xcolor}
\usepackage{caption}
\usepackage{float}
\floatstyle{plaintop}
\restylefloat{table}
\usepackage{subfigure} 
\usepackage{xcolor}
\usepackage{amsmath,amssymb}
\usepackage{algorithm} 
\usepackage{algorithmic} 
\usepackage[normalem]{ulem}

\newcommand{\ag}[1]{{\color{black} #1}}

\begin{document}
\title{JAMPR+/L2D: scalable neural heuristic for constrained vehicle routing problems in dynamic environment}
\titlerunning{Scalable Neural VRP Heuristics}

%
\author{Andrew Soroka\inst{1} \and
Alex Meshcheryakov\inst{1, 2} }
\authorrunning{A. Soroka et al.}
%
\institute{Moscow State University, Moscow, Russian Federation \email{andrew.soroka@student.msu.ru} \and Space Research Institute of RAS, Moscow, Russia \email{ mesch@cosmos.ru} }
\maketitle              
\begin{abstract}
The vehicle routing problems with real-world constraints (we consider vehicles capacity limits, time windows constrains, pickup-and-delivery multi-depo --- CPDPTW) pose significant computational challenges. While classical exact and heuristic methods remain effective to solve problems of small/medium size ($N\lesssim100$), they often lack adaptability and scalability for larger logistics tasks. In this work, we show how JAMPR+/L2D RL deep learning model,  proposed in to solve large CPDPTW problems can be adopted in the case of substantial changes of graph distance matrix. We test performance of JAMPR+/L2D model for medium-sized CVRP and VRPTW problems on CVRPLIB benchmarks: JAMPR+/L2D outperforms the state-of-the-art heuristic HGS in over 85\% of instances, achieving improvement in objective gap.  We show that the JAMPR+/L2D model trained on CPDPTW problem, generalizes well for tasks with simpler constraints (CVRP, VRPTW), for different problem sizes and for moderate changes in distance matrixes. For more substantial changes in distance matrixes, we propose here to make fast finetuning of JAMPR+: on ORTEC data (for CPDPTW) the proposed strategy remarkably reduces the objective gap without full model retraining, what will give both accuracy and rapid inference of the model in the practical routing scenarios with distance matrix changes.
\end{abstract}
\section{Introduction}
The Vehicle Routing Problem (VRP) is a core combinatorial optimization problem with wide impact in logistics, transportation, warehouse distribution, and network routing \cite{toth2014vehicle}. As a generalization of the Traveling Salesman Problem, VRP is NP-hard; exact algorithms have exponential worst-case complexity and quickly become impractical at scale, even on optimized systems \cite{dantzig1954solution,garey2002computers,kool2022deep}. Among practical variants, the Capacitated Pickup and Delivery Problem with Time Windows (CPDPTW) is particularly challenging, combining vehicle capacities, pickup–delivery precedence, and strict time windows that mirror real operational constraints in logistics and ride-sharing.

Classical heuristics and metaheuristics—Clarke–Wright savings, sweep, Genetic Algorithms, Simulated Annealing, Tabu Search—deliver good solutions under tight time budgets, but demand substantial expert tuning and often transfer poorly across instance distributions and constraint sets. This motivates \emph{neural heuristics}: data-driven policies that replace hand-designed rules with learned decision mechanisms.

Because routes are sequences, attention-based architectures (Transformers) from sequence modeling have proven effective for VRP variants, learning to attend to spatial and contextual node features and to construct feasible tours under complex constraints \cite{kool2018attention,lu2019learning}. The first attention-based VRP results around 2018 catalyzed a shift toward learned solvers \cite{vrp_overview}, while advances in GPT-like models suggest stronger generalization without explicit rule encoding \cite{deepseekai2024deepseekllmscalingopensource,openai2024gpt4technicalreport}. Building on this direction, we previously introduced a fully neural RL solver for large CPDPTW, JAMPR+/L2D, which extends JAMPR \cite{falkner2020learning} with L2D \cite{li2021learning} to improve scalability and produce fast suboptimal solutions suitable for industrial-scale deployments \cite{soroka2023solving,soroka_cpdptw_23}.

Despite notable progress, deployment remains difficult. Distribution shift—e.g., changes in distance-matrix statistics between training and inference—can degrade performance, and high-quality public datasets capturing real operational noise and constraints are scarce. Consequently, many methods are trained and validated on synthetic data that may not reflect real-world variability. This work aims at adaptive neural methods that remain robust to distribution shifts and handle the full CPDPTW complexity across scales and constraint combinations (including changes in distance matrices).

For evaluation, we use established benchmarks. CVRPLib \cite{uchoa2017new} aggregates diverse CVRP/VRPTW instances spanning capacities, customer distributions, and demand patterns. To assess realism, we also adopt the ORTEC dataset (EURO Meets NeurIPS 2022) \cite{ortec}, comprising operational scenarios with traffic-derived travel-time matrices, strict time windows, pickup–delivery constraints, multiple depots, and problem sizes from $\sim$200 to $>1000$ customers.

The paper is organized as follows. Chapter \ref{realted_Work} reviews exact, heuristic, and neural approaches for CPDPTW. Chapter \ref{Data} describes synthetic and real-world datasets. Chapter \ref{archs} presents the JAMPR+/L2D architecture. Chapter \ref{experiments} details training protocols and baselines. Chapter \ref{results} reports results and comparative analysis, followed by conclusions.

\section{Related Work} \label{realted_Work}
Among the various methods for solving vehicle routing problems, the exact approaches \cite{baldacci2012recent, costa2019exact} guarantee optimal solutions and have a strong theoretical basis. However, they struggle to efficiently handle the increasing complexity of modern vehicle routing issues.

In contrast, heuristics provide high-quality solutions at reasonable computational costs and demonstrate strong adaptability, making them popular among researchers and practitioners \cite{weinand2022research}. During the past few decades, significant effort has been made to develop vehicle routing heuristics, and according to \cite{braekers2016vehicle}, heuristics comprised more than 80\% of vehicle routing literature from 2009 to 2013.

Neural combinatorial optimization \cite{nazari2018reinforcement, kotary2021end} is a newer trend that quickly generates acceptable solutions using learned knowledge from large datasets, but its performance often falls short of advanced heuristics.

\subsection{Exact Methods}

Branch-and-bound (B\&B) is one of the most widely used exact optimization techniques for solving VRP and its many variants due to its generality and ability to guarantee optimal solutions. It systematically explores the solution space by partitioning it into smaller subproblems (branching) and using bounds to prune suboptimal regions, making it a natural fit for combinatorial problems like the VRP. Variants such as branch-and-cut and branch-and-price extend this framework to handle additional constraints efficiently. These methods are foundational in both academic and commercial solvers, including CPLEX and Gurobi, which rely heavily on B\&B for solving mixed-integer programming problems. Despite its theoretical elegance, the exponential worst-case complexity of B\&B severely limits its applicability to real-world problems, especially those with complex constraints like CPDPTW. In practice, it is primarily used for small-scale instances or as a benchmark for evaluating heuristics. Comprehensive overviews of B\&B and its role in VRP can be found in standard texts such as Combinatorial Optimization by Papadimitriou and Steiglitz, and Vehicle Routing: Problems, Methods, and Applications by Toth and Vigo \cite{toth2014vehicle}, as well as surveys like Braekers et al. \cite{braekers2016vehicle}.

\subsection{Heuristic Methods}

Research on vehicle routing heuristics has been ongoing since the VRP's inception. Heuristics are generally grouped into three main categories:

\begin{enumerate}
    \item Constructive heuristics: These methods build routing solutions from the ground up using empirical rules. They quickly produce feasible solutions and easily adapt to various variants of VRP. However, their solutions may not be close to optimal.\cite{soroka2023solving,soroka_cpdptw_23}
    \item Improvement heuristics: These methods enhance an existing solution through local search to find a local optimum. While efficient, they can be hindered by getting stuck in local optima, with the final solution's quality depending on the initial search conditions. \cite{soroka2023solving,soroka_cpdptw_23}
    \item Metaheuristics: Unlike constructive and improvement heuristics, metaheuristics utilize high-level algorithmic concepts. They draw inspiration from natural or physical phenomena to guide optimization strategies and offer efficient global search capabilities. \cite{holand1975adaptation, vidal2022hybrid, soroka2023solving, soroka_cpdptw_23}
\end{enumerate}

A full overview can be found in the authors' works \cite{soroka2023solving,soroka_cpdptw_23} or in Appendix \nameref{app_a}.

\subsection{Neural Heuristics}

The first deep learning model for solving VRP was introduced by Nazari et al. \cite{nazari2018reinforcement}, who adapted the Pointer Network (PtrNet) by Vinyals et al. \cite{vinyals2015pointer} for CVRP. Nazari et al. replaced the original model's RNN encoder with a linear layer sharing parameters. Later, Kool et al. \cite{kool2018attention} proposed the Attention Model (AM), substituting the architecture with a transformer model based on attention \cite{vaswani2017attention}. An advancement on this model is the JAMPR approach by Falkner et al. \cite{falkner2020learning}, which incorporates fully connected networks for the current path and truck positions, enabling the algorithm to tackle the CVRPTW problem.

Chen and Tian \cite{chen2019learning} introduced an RL-based method that selects a region on the graph and applies established local heuristics. Lu et al. \cite{lu2019learning} later refined this approach with a destruction operator. Li et al. \cite{li2021learning} proposed a deep learning approach to partitioning a set of points into subproblems and solving them using a black-box solver. The authors offered two methods: regression-based prediction of potential cost improvement and classification of the best subproblem. By simplifying the problem and employing classical metaheuristic methods in each subproblem, they achieved notable results on high-dimensional problems (over 1,000 points).

Soroka et al. \cite{soroka_cpdptw_23} proposed to use JAMPR+ deep reinforcement learning approach for solving the route optimization problems under realistic constraints (CPDPTW) of small/medium size. They demonstrated, that JAMPR+ neural heuristic solver outperform classical heuristics (OR-Tools) and also finds fast suboptimal solution. Furthermore, Soroka et al. \cite{soroka2023solving} proposed the L2D/JAMPR+L2D model, which solves large ($N\lesssim5000$) CPDPTW problems.

As conclusion for related literature review, we show in Table \ref{table_2} a comprehensive list of papers, where RL neuroheuristics were applied to VRP problems with various constraints. Columns in the Table represent different types of constraints, including: vehicles capacity, time windows of various kinds, multi-depo pickup and delivery constraints (PDP), problem size, changes in distance matrix statistical properties between training and test instances.  


\begin{table}[h]
\centering
\begin{tabular}{|l|l|lll|l|lll|l|}
\hline
\multirow{2}{*}{Paper}                                    & \multirow{2}{*}{Cap.} & \multicolumn{3}{c|}{Time Windows}                                                                                                                                                                         & \multirow{2}{*}{PDP} & \multicolumn{3}{c|}{Problem Size}                                                                                                                                                                                                            & \multirow{2}{*}{\begin{tabular}[c]{@{}l@{}}DMC\end{tabular}} \\ \cline{3-5} \cline{7-9}
                                                           &                       & \multicolumn{1}{l|}{\begin{tabular}[c]{@{}l@{}}TW1 \end{tabular}} & \multicolumn{1}{l|}{\begin{tabular}[c]{@{}l@{}}TW2 \end{tabular}} & \begin{tabular}[c]{@{}l@{}}TW3 
                                                           \end{tabular} 
                                                           &                      & \multicolumn{1}{l|}{\begin{tabular}[c]{@{}l@{}} \\ \textless{}100\end{tabular}} & \multicolumn{1}{l|}{\begin{tabular}[c]{@{}l@{}} \\ \textless{}1000\end{tabular}} & \begin{tabular}[c]{@{}l@{}} \\ \textgreater{}1000\end{tabular} &                                                                           \\ \hline
\begin{tabular}[c]{@{}l@{}}AM,  2019\cite{kool2018attention}\end{tabular}        & +                     & \multicolumn{1}{l|}{-}                                                   & \multicolumn{1}{l|}{-}                                                   & -                                                   & -                    & \multicolumn{1}{l|}{+}                                                               & \multicolumn{1}{l|}{-}                                                          & -                                                                   & -                                                                         \\ \hline
\begin{tabular}[c]{@{}l@{}}JAMPR,  2020\cite{falkner2020learning}\end{tabular}     & +                     & \multicolumn{1}{l|}{+}                                                   & \multicolumn{1}{l|}{+}                                                   & +                                                   & -                    & \multicolumn{1}{l|}{+}                                                               & \multicolumn{1}{l|}{-}                                                          & -                                                                   & -                                                                         \\ \hline
\begin{tabular}[c]{@{}l@{}}L2I,  2020\cite{lu2019learning}\end{tabular}       & +                     & \multicolumn{1}{l|}{-}                                                   & \multicolumn{1}{l|}{-}                                                   & -                                                   & -                    & \multicolumn{1}{l|}{+}                                                               & \multicolumn{1}{l|}{-}                                                          & -                                                                   & -                                                                         \\ \hline
\begin{tabular}[c]{@{}l@{}}W. Kool,  2021\cite{kool2022deep}\end{tabular}   & -                     & \multicolumn{1}{l|}{+}                                                   & \multicolumn{1}{l|}{+}                                                   & -                                                   & -                    & \multicolumn{1}{l|}{+}                                                               & \multicolumn{1}{l|}{-}                                                          & -                                                                   & -$^{\star}$                                                                        \\ \hline
\begin{tabular}[c]{@{}l@{}}L. Xin,  2021\cite{xin2021neurolkh}\end{tabular}    & +                     & \multicolumn{1}{l|}{+}                                                   & \multicolumn{1}{l|}{-}                                                   & -                                                   & +                    & \multicolumn{1}{l|}{+}                                                               & \multicolumn{1}{l|}{+}                                                          & -                                                                   & -                                                                         \\ \hline
\begin{tabular}[c]{@{}l@{}}Y. Ma,  2021\cite{ma2021learning}\end{tabular}     & +                     & \multicolumn{1}{l|}{-}                                                   & \multicolumn{1}{l|}{-}                                                   & -                                                   & -                    & \multicolumn{1}{l|}{+}                                                               & \multicolumn{1}{l|}{-}                                                          & -                                                                   & -                                                                         \\ \hline
\begin{tabular}[c]{@{}l@{}}C. Natalia, 2021\cite{natalia2021completion}\end{tabular} & +                     & \multicolumn{1}{l|}{-}                                                   & \multicolumn{1}{l|}{-}                                                   & +                                                   & -                    & \multicolumn{1}{l|}{+}                                                               & \multicolumn{1}{l|}{-}                                                          & -                                                                   & -                                                                         \\ \hline
\begin{tabular}[c]{@{}l@{}}L2D,  2022\cite{li2021learning}\end{tabular}       & +                     & \multicolumn{1}{l|}{+}                                                   & \multicolumn{1}{l|}{+}                                                   & +                                                   & -                    & \multicolumn{1}{l|}{-}                                                               & \multicolumn{1}{l|}{+}                                                          & +                                                                   & -$^{\star}$                                                                         \\ \hline
\begin{tabular}[c]{@{}l@{}}B. Rabecq, 2022\cite{rabecq2022deep}\end{tabular}  & +                     & \multicolumn{1}{l|}{-}                                                   & \multicolumn{1}{l|}{+}                                                   & +                                                   & +                    & \multicolumn{1}{l|}{+}                                                               & \multicolumn{1}{l|}{-}                                                          & -                                                                   & -                                                                         \\ \hline
\begin{tabular}[c]{@{}l@{}}Y. Ma,  2022\cite{ma2022efficient}\end{tabular}     & -                     & \multicolumn{1}{l|}{-}                                                   & \multicolumn{1}{l|}{-}                                                   & -                                                   & +                    & \multicolumn{1}{l|}{+}                                                               & \multicolumn{1}{l|}{-}                                                          & -                                                                   & -                                                                         \\ \hline
\begin{tabular}[c]{@{}l@{}}J. Choo,  2022\cite{choo2022simulation}\end{tabular}   & +                     & \multicolumn{1}{l|}{-}                                                   & \multicolumn{1}{l|}{-}                                                   & -                                                   & +                    & \multicolumn{1}{l|}{+}                                                               & \multicolumn{1}{l|}{+}                                                          & -                                                                   & -                                                                         \\ \hline
\begin{tabular}[c]{@{}l@{}}H.Cheng, 2023\cite{chen2022rule}\end{tabular}    & -                     & \multicolumn{1}{l|}{-}                                                   & \multicolumn{1}{l|}{-}                                                   & -                                                   & -                    & \multicolumn{1}{l|}{+}                                                               & \multicolumn{1}{l|}{+}                                                          & +                                                                   & -                                                                         \\ \hline
\begin{tabular}[c]{@{}l@{}} JAMPR+  2023\cite{soroka_cpdptw_23}\end{tabular}    & +                     & \multicolumn{1}{l|}{+}                                                   & \multicolumn{1}{l|}{-}                                                   & -                                                   & +                    & \multicolumn{1}{l|}{+}                                                               & \multicolumn{1}{l|}{+}                                                          & -                                                                   & -                                                                         \\ \hline
\begin{tabular}[c]{@{}l@{}} L2D/JAMPR+  2023\cite{soroka2023solving}\end{tabular}    & +                     & \multicolumn{1}{l|}{+}                                                   & \multicolumn{1}{l|}{-}                                                   & -                                                   & +                    & \multicolumn{1}{l|}{+}                                                               & \multicolumn{1}{l|}{+}                                                          & +                                                                   & -                                                                         \\ \hline
\begin{tabular}[c]{@{}l@{}} A. Hottung  2024\cite{hottung2024polynet}\end{tabular}    & +                     & \multicolumn{1}{l|}{+}                                                   & \multicolumn{1}{l|}{-}                                                   & -                                                   & -                    & \multicolumn{1}{l|}{+}                                                               & \multicolumn{1}{l|}{+}                                                          & -                                                                   & -                                                                         \\ \hline
\begin{tabular}[c]{@{}l@{}}G. Santiyuda 2024\cite{santiyuda2024multi}\end{tabular}    & -                     & \multicolumn{1}{l|}{+}                                                   & \multicolumn{1}{l|}{-}                                                   & -                                                   & +                    & \multicolumn{1}{l|}{+}                                                               & \multicolumn{1}{l|}{-}                                                          & -                                                                   & -                                                                         \\ \hline
\begin{tabular}[c]{@{}l@{}}J. Fitzpatrick 2024\cite{fitzpatrick2024scalable}\end{tabular}    & +                     & \multicolumn{1}{l|}{-}                                                   & \multicolumn{1}{l|}{-}                                                   & -                                                   & -                    & \multicolumn{1}{l|}{+}                                                               & \multicolumn{1}{l|}{+}                                                          & +                                                                   & -                                                                         \\ \hline
\begin{tabular}[c]{@{}l@{}}A. Hottung 2025\cite{hottung2025neural}\end{tabular}    & +                     & \multicolumn{1}{l|}{+}                                                   & \multicolumn{1}{l|}{-}                                                   & -                                                   & -                    & \multicolumn{1}{l|}{+}                                                               & \multicolumn{1}{l|}{+}                                                          & +                                                                   & -                                                                         \\ \hline
\begin{tabular}[c]{@{}l@{}}Y. Wang 2025\cite{wang2025multi}\end{tabular}    & +                     & \multicolumn{1}{l|}{+}                                                   & \multicolumn{1}{l|}{-}                                                   & -                                                   & +                    & \multicolumn{1}{l|}{+}                                                               & \multicolumn{1}{l|}{-}                                                          & -                                                                   & -                                                                         \\ \hline
\begin{tabular}[c]{@{}l@{}}This work \end{tabular}    & +                     & \multicolumn{1}{l|}{+}                                                   & \multicolumn{1}{l|}{+$^{\star\star}$}                                                   & +$^{\star\star}$                                                   & +                    & \multicolumn{1}{l|}{+}                                                               & \multicolumn{1}{l|}{+}                                                          & +                                                                   & +                                                                         \\ \hline
\end{tabular}

$^{\star}$ --- proposed models were fount not robust against distribution changes. \\
$^{\star\star}$ --- we tested simpler time constraints and model works well on them. (see \S\ref{tw_changes}) \\

\caption{List of papers, where RL neuroheuristics were applied to constrained VRP problems. Time windows constraints are considered of three types: hard (TW1), medium (TW2) and soft (TW3). The last column (DMC) means Distance Matrix Changes. "+" against method denotes the presence of a constraint in the corresponding paper, minus denotes its absence or unsatisfactory result. }
\label{table_2}
\end{table}

\section{Data} \label{Data}

The algorithms considered in this research aim to solve the CPDPTW problem and more simpler problems (like CVRP, VRPTW etc), focusing on how they handle multiple constraints simultaneously. Not all classical algorithms can manage these constraints effectively. Moreover, there has not been an academic proposal for the implementation of the HGS algorithm in a CPDPTW solver to date. Since HGS remains the state-of-the-art approach, we use the VRPTW problem as a reference, as it contains the strongest constraint among the remaining constraints.

In this work we will use 3 different datasets in order to train and test our model: (i) CVRPLib, (ii) standard generated instances (according to Solomon reference set), (iii) real-world instances from Neuro IPS VRP competition. Below we describe these data in detail.

{\bf CVRPLib.} CVRPLib is a well-established benchmark library designed to support the development and evaluation of algorithms for the Capacitated Vehicle Routing Problem (CVRP) and its common variants. Introduced by Uchoa et al. \cite{uchoa2017new}, the library consolidates a diverse set of instances collected from academic literature, restructured under a unified format to facilitate reproducibility and fair comparison across methods. CVRPLib includes problems of varying sizes, from small instances with fewer than 20 customers to large-scale problems with over 1000 nodes. These instances incorporate a range of characteristics, including different vehicle capacities, customer distributions, and demand patterns, making the dataset a valuable benchmark for testing algorithm scalability and robustness. While CVRPLib does not include complex real-world constraints pickup-and-delivery precedence, it remains one of the most widely used resources for validating heuristic and exact solvers in controlled experimental settings. In our work, we adopt CVRPLib as a baseline dataset to benchmark our approach against established methods, leveraging its wide recognition and consistent structure for standardized evaluation.

{\bf Solomon.} For the generated samples, we selected suitable instances from the well-known reference set by \textit{Solomon}, based on the R201 statistics \cite{solomon1987algorithms}. The truck volumes were set as $Q_{50} = 750$ and $Q_{100} = 1000$ for problem size 200, respectively. The total time horizon is defined as $[a_0, b_0]$, with $a_0 = 0$ for all examples, and the right boundary $b_0$ varying depending on the problem size, specifically 1000 for 200 points. Additionally, the service duration $h_i$ for each point is uniformly set to 10.

{\bf ORTEC.} For real-world samples, we used data from the Neuro IPS VRP competition. The VRPTW problem comprises \( n \) locations (a depot and \( n-1 \) customers), a distance matrix \( D \) specifying travel times between locations, customer demands \( q_i \), and vehicle capacity \( Q \). Each location \( i \) has a service time \( s_i \) and a time window \([t_i, T_i]\) within which delivery must start. Vehicles can arrive early but must wait for the time window to open; late deliveries are not allowed. The challenge treats time-window constraints as strict.

We also utilized real instances from ORTEC \cite{ortec} with explicit duration matrices, providing real-time travel information between customers. The number of customers ranges from 200 to 1000. Each customer has coordinates, demand, a time window, and service duration, with service starting within the time window. Vehicles must adhere to customer time windows, and total demand served by each vehicle must not exceed its capacity. The goal is to minimize total driving time, excluding wait times. There are no practical limits or objectives regarding the number of vehicles used.

To solve the fixed-size problem, we sampled clients from ORTEC \cite{ortec} tasks while retaining the original fixed size. We also saved the distance matrix and goods volumes from the original problem, along with time windows. For additional fine-tuning, we sampled 1,000,000 instances from the competition training set and 1,000 instances for testing from the competition test set for 200-sized problems.

\section{JAMPR+/L2D model and comparison with other VRP RL architectures} 
\label{archs}

There are many works describing approaches to solving route optimization problems using neural networks\cite{li2021learning, bi2022learning, zhang2023review}. The most popular constraints --- capacity and time windows were first successfully added in AM\cite{kool2018attention} and JAMPR\cite{falkner2020learning} using the idea of sequentially constructing a route point by point, using the basis of a transformer: encoder and decoder. In the first case, the capacity was simply added to the Node Encoder, the second paper is a successor to the first, but the processing time windows are assigned to additional encoders. We tested how such approaches work if we want to add a new constraint, in our case --- multi-depot (PDP). We separate these two approaches, since they are conceptually diametrically opposed: what will bring more benefit, mixing all input constraints into one encoder and hoping that the network itself will learn the necessary heuristics, or manually feeding it with additional encoders to handle the conceptual parts of the problem (tour and vehicles)?

\begin{figure}[!ht]
\centering

\begin{minipage}[t]{.5\textwidth}
  \centering
  \includegraphics[width=\linewidth]{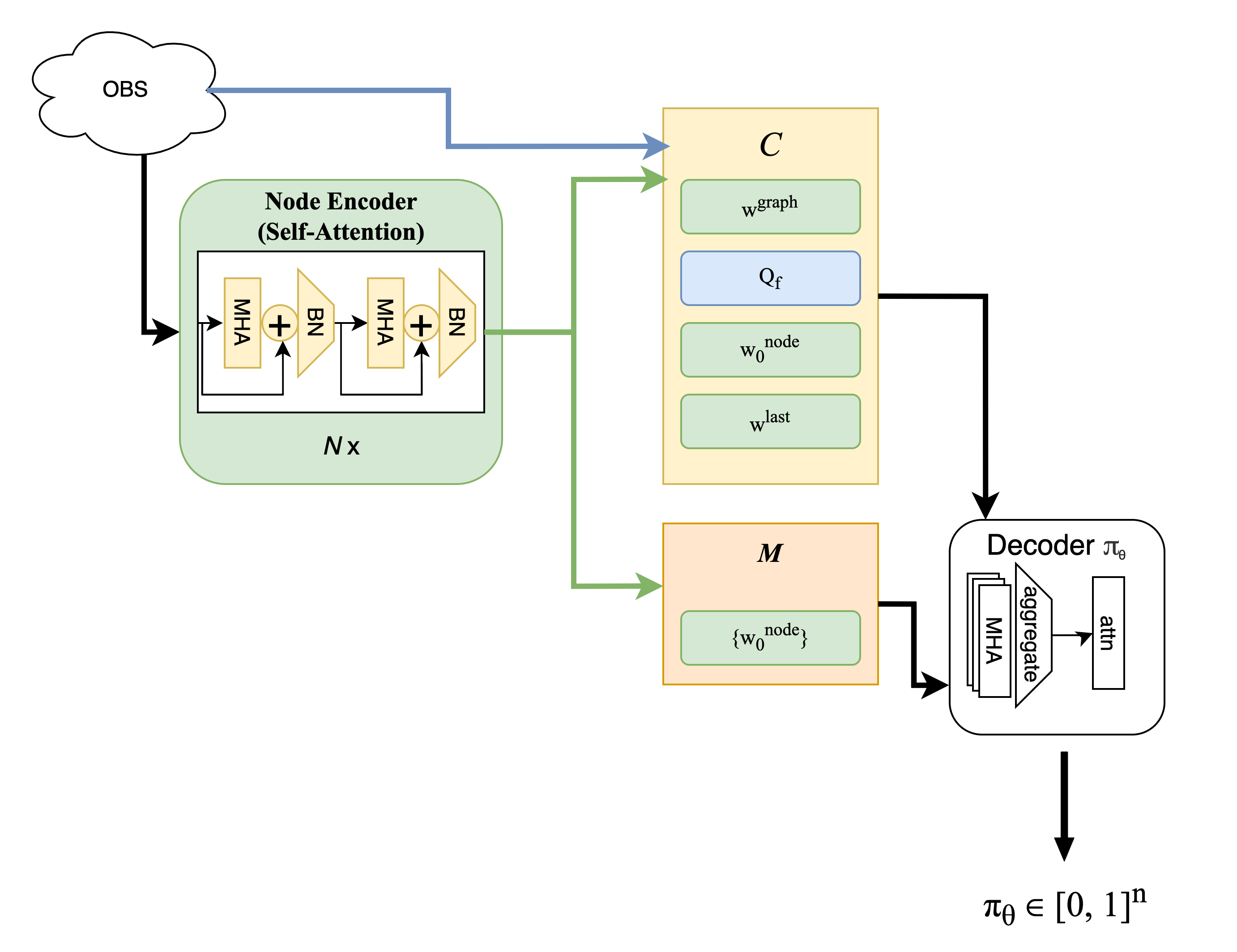}
  \vspace{2pt}\par AM (CVRP)
\end{minipage}\hfill
\begin{minipage}[t]{.5\textwidth}
  \centering
  \includegraphics[width=\linewidth]{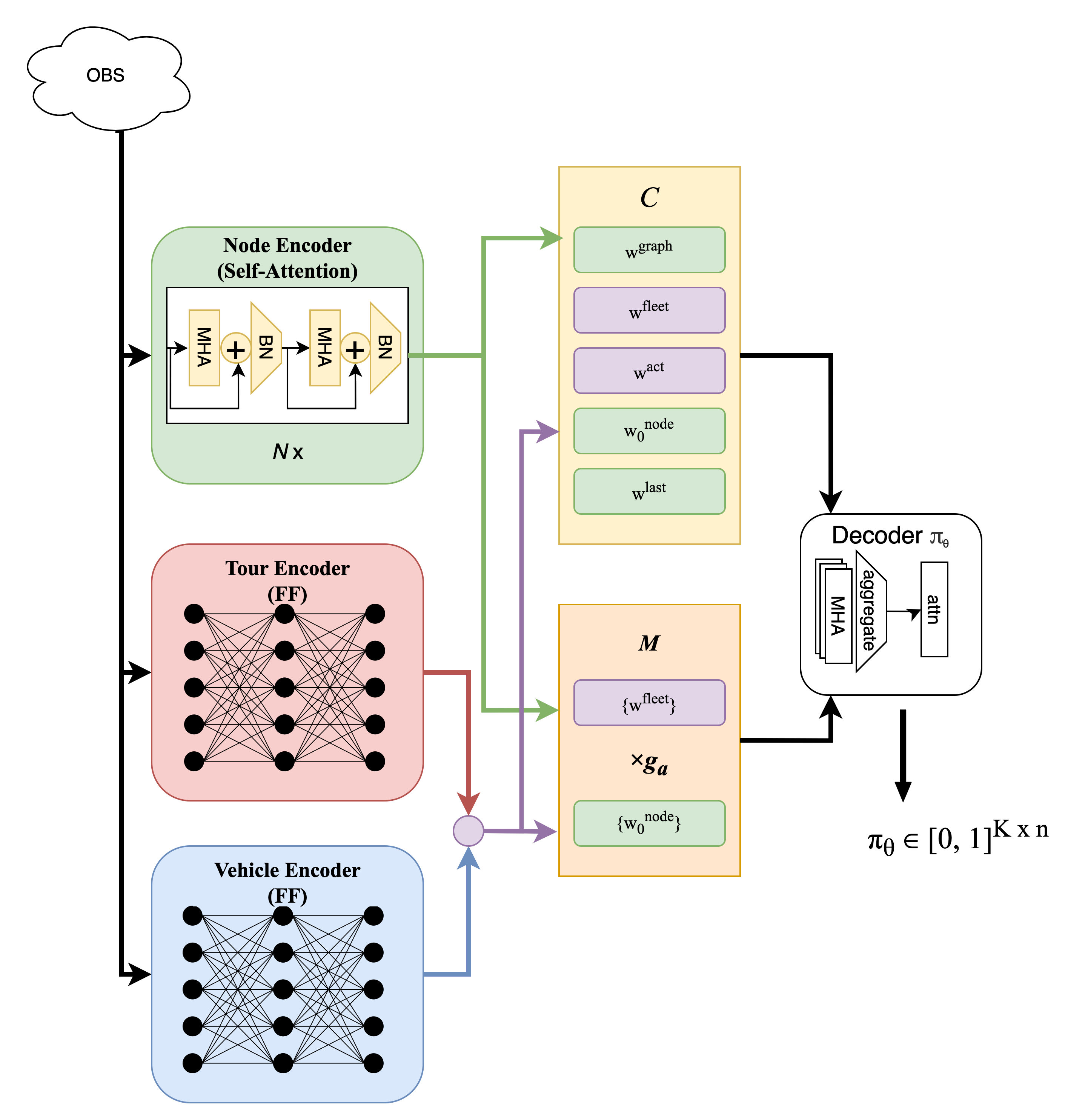}
  \vspace{2pt}\par JAMPR (CVRPTW)
\end{minipage}

\vspace{0.8em}

\begin{minipage}[t]{.6\textwidth}
  \centering
  \includegraphics[width=\linewidth]{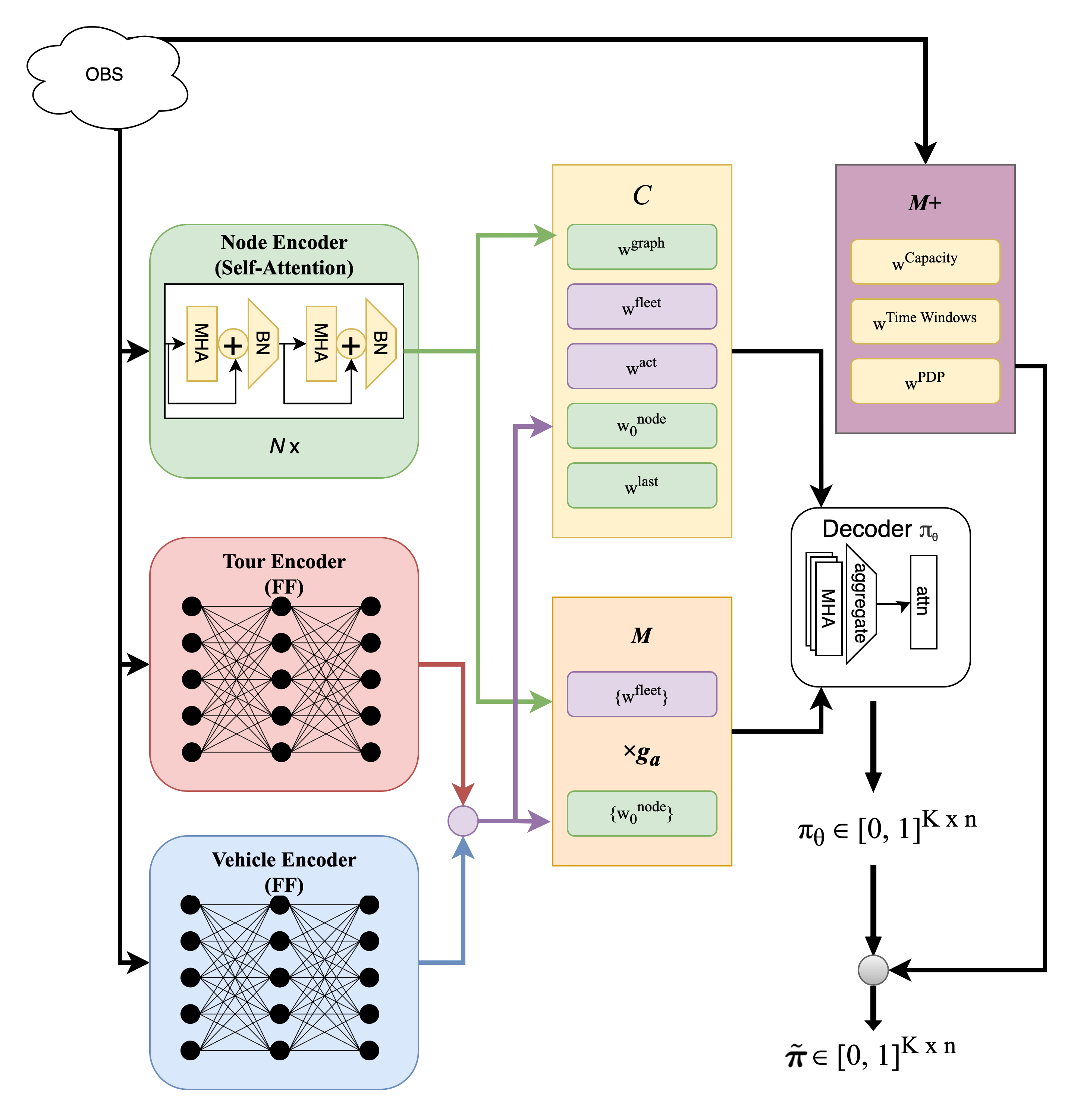}
  \vspace{2pt}\par JAMPR+ (CPDPTW)
\end{minipage}\hfill
\begin{minipage}[t]{.4\textwidth}
  \centering
  \includegraphics[width=\linewidth]{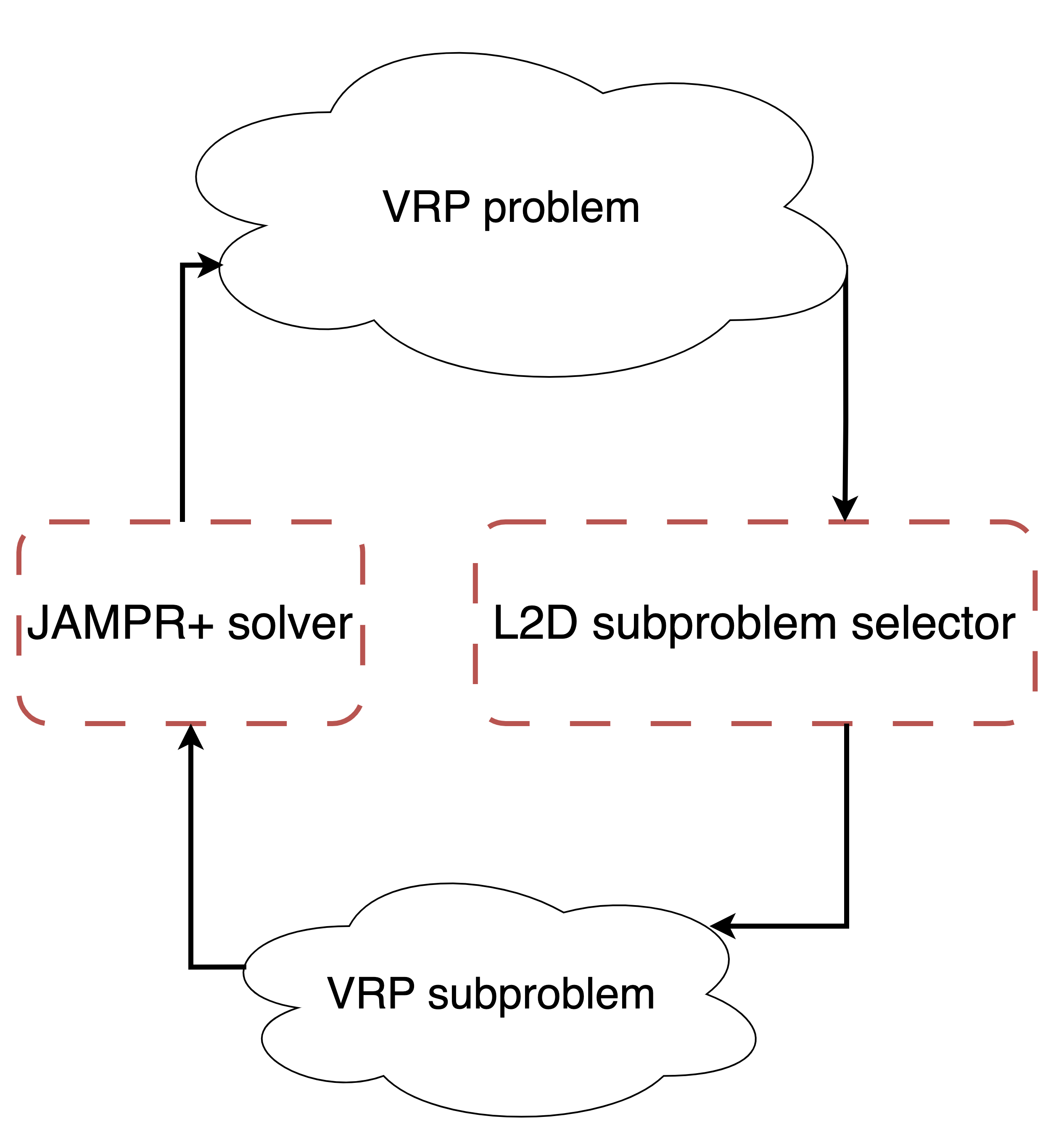}
  \vspace{2pt}\par JAMPR+/L2D (CPDPTW)
\end{minipage}

\caption{Deep RL transformer architectures for VRP problems with different constraints. Top row: AM (left), JAMPR (right). Bottom row: JAMPR+ (left), JAMPR+/L2D (right).}
\label{fig:architecture_selection}
\end{figure}

\begin{figure}[!h]
\centering

\begin{minipage}[t]{.483\textwidth}
  \centering
  \includegraphics[width=\linewidth]{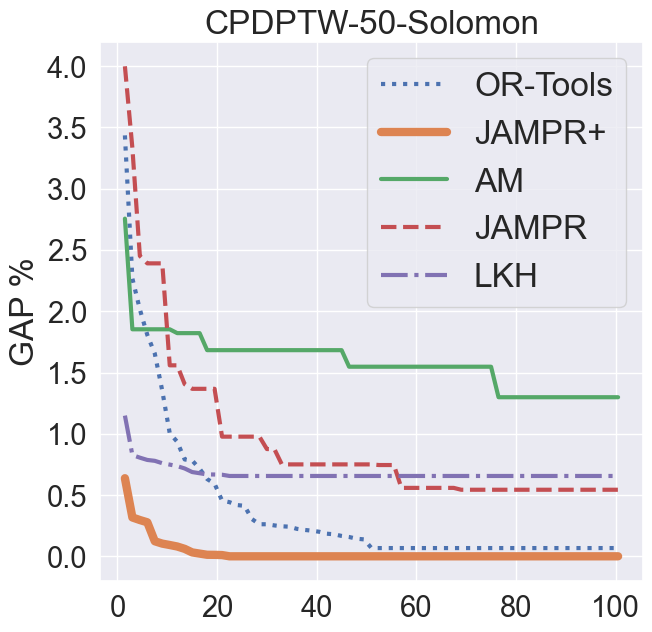}
\end{minipage}\hfill
\begin{minipage}[t]{.457\textwidth}
  \centering
  \includegraphics[width=\linewidth]{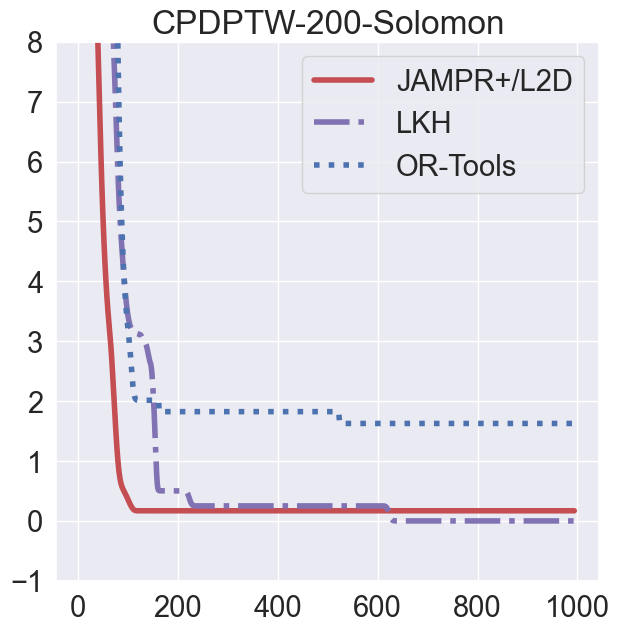}
\end{minipage}

\vspace{0.8em}

\begin{minipage}[t]{.491\textwidth}
  \centering
  \includegraphics[width=\linewidth]{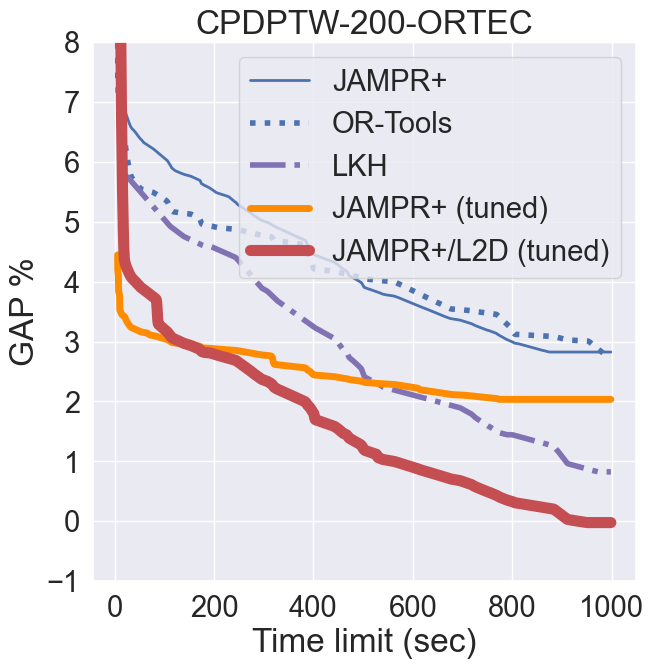}
\end{minipage}\hfill
\begin{minipage}[t]{.449\textwidth}
  \centering
  \includegraphics[width=\linewidth]{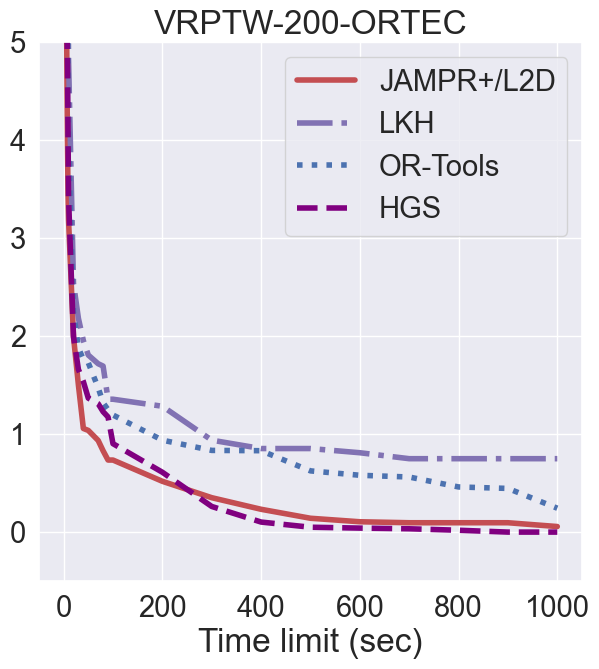}
\end{minipage}

\caption{Solver performance over runtime: curves show the \textbf{mean} GAP (\%, lower is better) versus time limit, where GAP is computed relative to the best solution observed (BSO) across all methods and all time points (global best across curves). Top panels represent generated Solomon distribution. Bottom panels represent real-world ORTEC data. Left upper panel: JAMPR+ best basic solver across all heuristic models for CPDPTW with size 50. Right upper: JAMPR+/L2D outperform heuristic without finetuning for larger problem size on known distance matrix distribution for CPDPTW with size 200. Left bottom: tuned mask of JAMPR+/L2D gains best solution for unseen distance matrix distribution over all heuristics for CPDPTW with size 200. Right bottom:  JAMPR+/L2D outperforms all heuristic on weaker constraints without retraining for VRPTW with size 200. Proposed JAMPR+/L2D model is the best algorithm for solving route optimization problems with sizes 50 and 200 and VRPTW and CPDPTW constraints.}
\label{fig:architecture_quality_comparison}
\label{fig:first_comparison}
\end{figure}

AM and JAMPR approaches are well-known, while adding a new constraint to their pipeline is not obvious: it requires extending the heavyweight Node Encoder, which affects the speed of operation. We propose lightweight trainable masks for each new constraint. We tested three architectures shown in the figure \ref{fig:architecture_selection}. We trained all architectures same number of epochs (723) on Solomon \cite{solomon1987algorithms} distribution and tested on the same testing subset. \ag{Left upper panel on figure \ref{fig:architecture_quality_comparison} show on Solomon-50 distribution JAMPR+ converges faster and attains a lower GAP ($\approx$ 0.5–4.0 pp improvement during evaluation time) than AM, JAMPR, and OR-Tools.}

In our previous work we showed that best choice for large sized tasks is JAMPR+/L2D \cite{soroka2023solving}. The main idea of this approach is to use the divide and conquer technique: a neural network breaks a problem into subproblems that are quickly and optimally solved by an algorithm - a black box. We modified the black box proposed in original work \cite{li2021learning} by replacing it with our JAMPR+ model, trained to solve problems of size 50. Thus, the ensemble of these two models allowed us to solve problems of any dimension, preserving the suboptimal behavior of neural networks. Size 50 choosen since we cannot train pure JAMPR+ enough: time increases with size of problems. \ag{In the figure \ref{fig:architecture_quality_comparison} left bottom panel we show how  on ORTEC-200 distribution, JAMPR+/L2D yields the lowest GAP across the entire time range, outperforming JAMPR+ (vanilla), JAMPR+ with tuned masks, LKH, and OR-Tools.}


We have obtained a robust model that provides a fast optimal solution for problems of different sizes. Mainly using our proposed approach with trainable masks. As a result of this chapter, we offer a guide to using neural network approaches to solve problems of different dimensions: for problems less than 50, we recommend choosing pure JAMPR+ with masks, for problems of medium and high dimensions, the best option is to use JAMPR+/L2D.

We used JAMPR+ from our work \cite{soroka_cpdptw_23} as the basic neural network algorithm. The architecture can be found in the figure \ref{fig:architecture_selection}. In this approach, learned masks (M+) are used to modify the model's output policies, enabling it to better accommodate different sets of constraints. Each mask has size as policy tensor ($[n \times n ]$, where n is the size of the problem). Typically, training a neural heuristics model involves training the entire network, which requires substantial training time. This tradeoff might be acceptable when working with generated data, as its volume is practically limitless, but it becomes impractical when dealing with real data.

Our approach is to train the network on generated data to achieve a satisfactory baseline quality and then fine-tune only the policy mask using the target data. By freezing the network weights and training just the mask, we significantly reduced the training time from 14 days to just 10 minutes for a problem size of 200. 

We used JAMPR+/L2D model \cite{soroka2023solving} trained on artificial data to solve the CPDPTW large-sized problem. The architecture of JAMPR+/L2D pipeline an be found in figure \ref{fig:architecture_selection} bottom panel. The description of L2D network can be found in original work \cite{li2021learning}.

\section{Experiments} \label{experiments}

We selected 1,000,000 samples of real training data for a one-time pass through the data set. This further trained the model to solve the CPDPTW problem.

For problems with fewer constraints, model inference is performed by adjusting the mask to match the available constraints. Inference and training occur on the same hardware: 16 x AMD EPYC 7763 64-Core Processors, 1 x NVIDIA A100-SXM4-80GB GPU, and 64 GB RAM.

We considered three classical heuristic approaches as baselines: OR-Tools, LKH, and HGS. Note that HGS has not been proposed for CPDPTW, so we compared it with the VRPTW problem. For other approaches, we present comparisons on the CPDPTW problem. As a route cost metric, we use the total path length required to deliver all goods. We evaluate the best average path cost offered by each of the considered algorithms within the allotted optimization time. The intervals of 70 and 1000 seconds of optimization are of particular interest. In the first case, we examine a fast suboptimal solution, while in the second, we explore the theoretical quality of the solution under more significant restrictions on optimization time.

\section{Results} \label{results}

This section presents a comprehensive evaluation of our proposed neural routing approach under varying problem constraints, instance types, and data distributions. In Section \ref{archs}, we compare three deep reinforcement learning architectures—AM, JAMPR, and our improved JAMPR+/L2D on synthetic data, highlighting the impact of architectural design and constraint handling mechanisms on optimization performance. Section \ref{benchs} evaluates JAMPR+/L2D on standard benchmark datasets (CVRP and VRPTW) from CVRPLib, comparing its accuracy and robustness to state-of-the-art heuristics, including HGS, LKH, and OR-Tools. We provide quantitative GAP analyses and show that JAMPR+/L2D outperforms HGS in approximately 85\% of VRPTW instances. In Section \ref{real_world_data}, we study the model’s sensitivity to distributional shifts by testing on real-world ORTEC data. We demonstrate that although JAMPR+/L2D exhibits performance degradation on out-of-distribution data, our proposed mask-based finetuning strategy restores inference quality and efficiency, reducing GAP from 2\% to 0\% without full retraining.

\subsection{Results on CVRP and VRPTW benchmarks} \label{benchs}

We want to test how applicable our model is to lighter constraints, and whether the absence or presence of additional constraints affects its robustness. Relaxing the constraints also allows us to compare the quality of our predictions with the HGS heuristic, which is considered by some studies to be the SoTA solution in the field. We evaluated classical benchmarks with constraints on capacity and time-windows. It is important to note that currently, there is no publicly available implementation of HGS capable of addressing all real-world constraints. HGS algorithms for CVRP and VRPTW are also implemented separately.

\subsubsection{Time windows type changes} \label{tw_changes}

Our first assumption was to test how much the complexity of time windows affects the change in the quality of the model predictions. We had a model trained on time windows of a difficult type: missing a time slot means an unsolved problem. We launched the inference of this model on two other types of time windows: soft (you can arrive before the time slot and wait, as well as after and get a fine), medium (you can arrive before the time slot and wait). The results were the same in all three cases; completely neural approaches bypass classical heuristic ones in the first seconds of optimization, giving a quick solution, gradually asymptotically converging on the exact result.

\subsubsection{CVRP}

\begin{figure}[!ht]
    \centering
    \includegraphics[width=0.3\textwidth]{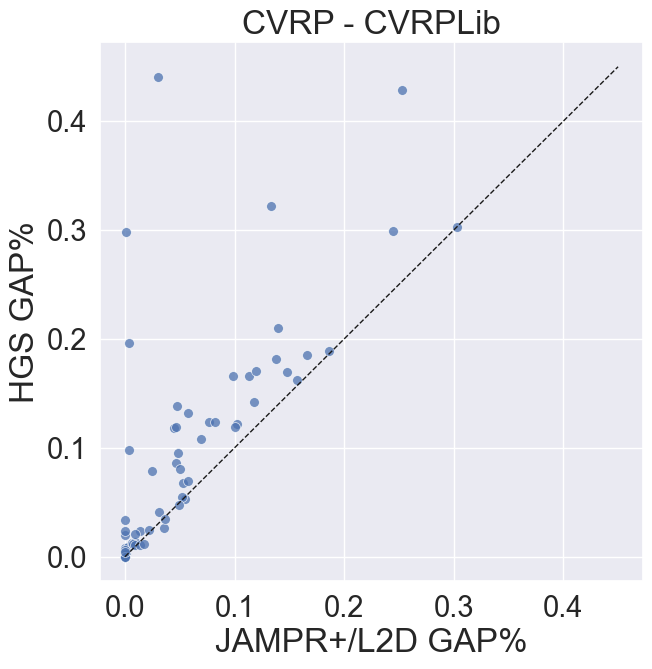}
    \includegraphics[width=0.3\textwidth]{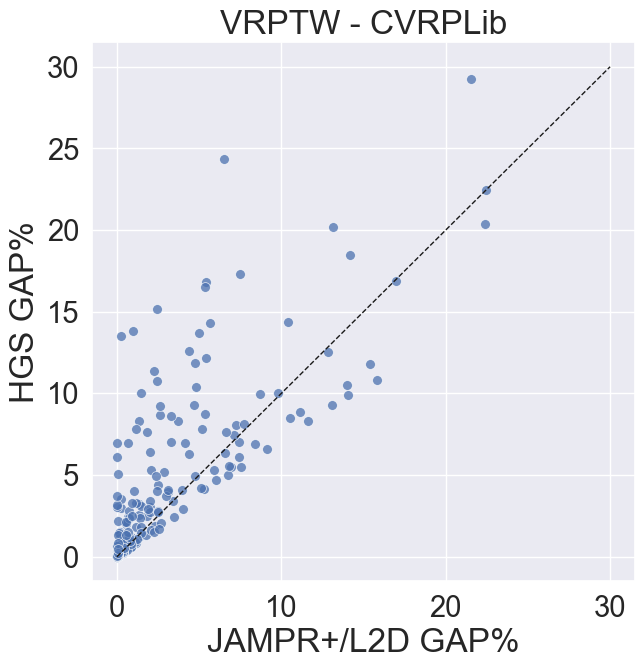}
    \includegraphics[width=0.3\textwidth]{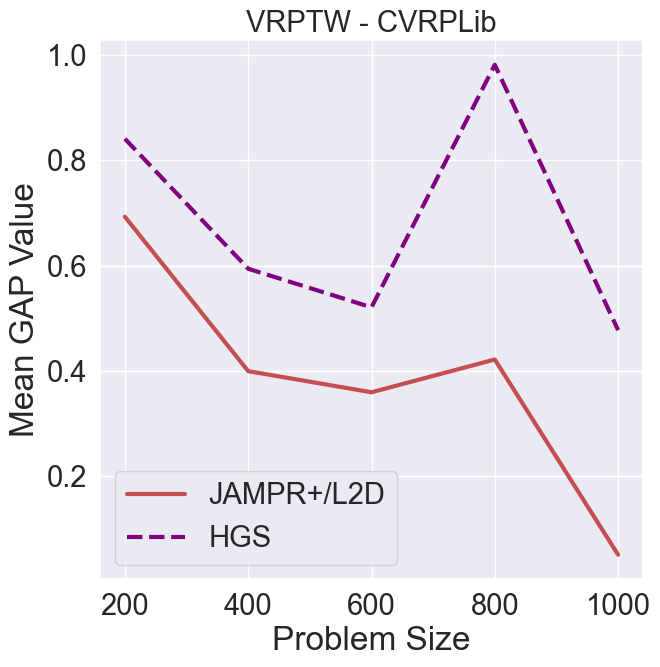}
\caption{GAPs(\%) for JAMPR+/L2D and HGS on CVRPLib test instances relative to best-known solutions (BKS). JAMPR+/L2D outperforms HGS on CVRP (left panel) and VRPTW (central panel) constraints and on problem sizes from 200 to 1000 (right panel). See text for details.}
   
\label{fig:cvrp_uchoa_jampr_hgs} 
\end{figure}

For comparisons on the problem with load volume constraints, we utilized the Uchoa dataset. The baseline is BKS (Best Known Solutions)—the best solutions known to date. It is worth mentioning that no single algorithm consistently provides the best solution for all instances within a dataset. Typically, various exact approaches are required, most of which demand extensive computational time. Therefore, we also included results from heuristic approaches such as HGS, LKH, and OR-Tools. \ag{The graph below shows the distribution of GAP values for all problem instances within the Uchoa dataset (figure \ref{fig:cvrp_uchoa_jampr_hgs}, left panel, each point is an instance). Most points lie above the $y{=}x$ diagonal, indicating HGS has a larger GAP than JAMPR+/L2D, with errors clustered near zero. }

We conclude that the proposed JAMPR+/L2D approach ideally solves problems from the Uchoa set. Thus, the use of neural network heuristics is justified for solving optimization problems with CVRP constraints.

\subsubsection{VRPTW}

As in the CVRP problem, we compared our results with BKS for the VRPTW problem. The figure \ref{fig:cvrp_uchoa_jampr_hgs}, central panel illustrates the GAP distributions between HGS and BKS and between  JAMPR+/L2D and BKS: the solutions almost identically show a near-zero GAP distribution relative to BKS. This suggests that only a few examples are able to lead the models to suboptimal solutions. In order to examine in more detail the differences in the GAP distributions between the two algorithms under consideration, we constructed a Scatterplot between the GAPs of HGS and BKS and between JAMPR+/L2D and BKS. \ag{The figure \ref{fig:cvrp_uchoa_jampr_hgs}, central panel shows that for the overwhelming majority of erroneous examples, HGS makes more mistakes than JAMPR+/L2D. The maximum lag in GAP for JAMPR+/L2D is no more than 2\%, while for HGS relative to JAMPR+/L2D it reaches 18\%.}

In the figure \ref{fig:cvrp_uchoa_jampr_hgs}, right panel we showed how the average GAP indicator depends on the problem size for the JAMPR+/L2D and HGS algorithms relative to BKS. It can be seen that JAMPR+/L2D consistently has an optimal metric value of up to 1-2\% regardless of the problem size. \ag{JAMPR+/L2D stays below HGS across all sizes and the advantage widens for larger $n$ (notably around $n\approx800$--$1000$), while HGS is non-monotonic.}

All this suggests that JAMPR+/L2D outperforms classical heuristic approaches on problems with similar constraints, being a priority algorithm for their optimization.

\subsection{CPDPTW results on ORTEC real-world data with distribution changes} \label{real_world_data}

Evaluating routing models on real-world data is a critical step toward their practical deployment. While synthetic benchmarks are useful for development and controlled comparison, they often fail to reflect the variability, noise, and complexity of real logistics environments. Therefore, one of the main goals of this study is to assess how well our proposed JAMPR+/L2D model generalizes to real-world instances, using the ORTEC dataset introduced in the EURO Meets NeurIPS 2022 competition \cite{ortec}.

Figure~\ref{fig:first_comparison} right upper panel presents the performance of different heuristics on the CPDPTW-200 problem, comparing optimization time and average percentage GAP from the best-known solution. In left panel, the model is tested on data drawn from the same distribution it was trained on. JAMPR+/L2D starts with a higher initial GAP but quickly converges, ultimately reaching a solution quality close to LKH and within 2\% GAP compared to OR-Tools.

Figure~\ref{fig:first_comparison}, right bottom panel shows that after mask finetuning, JAMPR+/L2D regains its ability to quickly find high-quality suboptimal solutions on ORTEC data, matching the asymptotic performance of state-of-the-art heuristics such as HGS and maintaining the neural model's characteristic fast response in the first seconds of optimization. This highlights domain-shift effects on the vanilla model and how mask finetuning restores fast convergence and closes the quality gap.

\textbf{Summary.} This experiment confirms that neural heuristics like JAMPR+/L2D are vulnerable to distribution shifts when applied to realistic data. However, the proposed mask finetuning method successfully recovers both inference quality and the model’s ability to generate fast suboptimal solutions. We recommend this approach as an effective strategy for adapting pretrained neural solvers to real-world CPDPTW problems with complex and evolving constraints.

\section{Conclusions} \label{conclusions}

In this paper, we propose a neural network solution JAMPR+/L2D for the Capacitated Pickup and Delivery Problem with Time Windows (CPDPTW), a challenging variant of the Vehicle Routing Problem that reflects real-world constraints. We introduce lightweight trainable policy masks for JAMPR+/L2D to enable fast adaptation to different constraint combinations(CPDPTW, VRPTW, CVRP) and distance matrices distance changes without full retraining full weights. 

We draw the following conclusions:
\begin{enumerate}
    \item We show that our JAMPR+ approach (with trainable masks) outperforms AM and JAMPR as basic solver for CPDPTW (problem size 50). JAMPR+/L2D model with trainable masks (CPDPTW problem size 200) is computationally efficient than metaheuristics (LKH, OR-Tools)
    
    \item We benchmarked JAMPR+/L2D on standard datasets with lighter constraints (CVRP, VRPTW) from the well-known CVRPLib \cite{uchoa2017new}. On CVRP tasks, JAMPR+/L2D achieved the best known solutions. For VRPTW, our model gained up to 2\% improvement in mean GAP over metaheuristics: HGS, LKH, OR-Tools. JAMPR+/L2D outperforms HGS in $\approx$ 85\% of instances for VRPTW (problem size 200 --- 1000).
    \item We benchmarked JAMPR+/L2D on real-world ORTEC data for CPDPTW constraints. The mask finetuning technique  (that updates only the constraint-specific masks without altering the pretrained model weights) allows our model outperform heuristics.
\end{enumerate}

Our approach aims to combine the flexibility of neural network models with the quality of metaheuristics, enabling practical deployment in logistics systems.

\bibliographystyle{splncs04}
\bibliography{mybibliography}

@article{vidal2022hybrid,
  title={Hybrid genetic search for the CVRP: Open-source implementation and SWAP* Neighborhood},
  author={Vidal, Thibaut},
  journal={Computers \& Operations Research},
  volume={140},
  pages={105643},
  year={2022},
  publisher={Elsevier}
}

@article{li2021learning,
  title={Learning to delegate for large-scale vehicle routing},
  author={Li, Sirui and Yan, Zhongxia and Wu, Cathy},
  journal={Advances in Neural Information Processing Systems},
  volume={34},
  year={2021}
}

@article{vaswani2017attention,
  title={Attention is all you need},
  author={Vaswani, Ashish and Shazeer, Noam and Parmar, Niki and Uszkoreit, Jakob and Jones, Llion and Gomez, Aidan N and Kaiser, {\L}ukasz and Polosukhin, Illia},
  journal={Advances in neural information processing systems},
  volume={30},
  year={2017}
}

@article{falkner2020learning,
  title={Learning to solve vehicle routing problems with time windows through joint attention},
  author={Falkner, Jonas K and Schmidt-Thieme, Lars},
  journal={arXiv preprint arXiv:2006.09100},
  year={2020}
}

@article{kool2018attention,
  title={Attention, learn to solve routing problems!},
  author={Kool, Wouter and Van Hoof, Herke and Welling, Max},
  journal={arXiv preprint arXiv:1803.08475},
  year={2018}
}

@article{solomon1987algorithms,
  title={Algorithms for the vehicle routing and scheduling problems with time window constraints},
  author={Solomon, Marius M},
  journal={Operations research},
  volume={35},
  number={2},
  pages={254--265},
  year={1987},
  publisher={Informs}
}

@inproceedings{lu2019learning,
  title={A learning-based iterative method for solving vehicle routing problems},
  author={Lu, Hao and Zhang, Xingwen and Yang, Shuang},
  booktitle={International conference on learning representations},
  year={2019}
}

@article{nazari2018reinforcement,
  title={Reinforcement learning for solving the vehicle routing problem},
  author={Nazari, Mohammadreza and Oroojlooy, Afshin and Snyder, Lawrence and Tak{\'a}c, Martin},
  journal={Advances in neural information processing systems},
  volume={31},
  year={2018}
}

@article{vinyals2015pointer,
  title={Pointer networks},
  author={Vinyals, Oriol and Fortunato, Meire and Jaitly, Navdeep},
  journal={Advances in neural information processing systems},
  volume={28},
  year={2015}
}

@article{chen2019learning,
  title={Learning to perform local rewriting for combinatorial optimization},
  author={Chen, Xinyun and Tian, Yuandong},
  journal={Advances in Neural Information Processing Systems},
  volume={32},
  year={2019}
}

@article{dantzig1954solution,
  title={Solution of a large-scale traveling-salesman problem},
  author={Dantzig, George and Fulkerson, Ray and Johnson, Selmer},
  journal={Journal of the operations research society of America},
  volume={2},
  number={4},
  pages={393--410},
  year={1954},
  publisher={INFORMS}
}

@inproceedings{kool2022deep,
  title={Deep policy dynamic programming for vehicle routing problems},
  author={Kool, Wouter and van Hoof, Herke and Gromicho, Joaquim and Welling, Max},
  booktitle={International conference on integration of constraint programming, artificial intelligence, and operations research},
  pages={190--213},
  year={2022},
  organization={Springer}
}

@article{braekers2016vehicle,
  title={The vehicle routing problem: State of the art classification and review},
  author={Braekers, Kris and Ramaekers, Katrien and Van Nieuwenhuyse, Inneke},
  journal={Computers \& Industrial Engineering},
  volume={99},
  pages={300--313},
  year={2016},
  publisher={Elsevier}
}

@article{vrp_overview,
  title={A bibliometric visualized analysis and classification of vehicle routing problem research},
  author={Ni, Qiuping and Tang, Yuanxiang},
  journal={Sustainability},
  volume={15},
  number={9},
  pages={7394},
  year={2023},
  publisher={MDPI}
}

@ARTICLE{soroka_cpdptw_23,
    author = {Soroka, A. G. and Meshcheryakov, A. V. and Gerasimov, S. V.},
    title = {Deep Reinforcement Learning for the Capacitated Pickup and Delivery Problem with Time Windows},
    journal = {Pattern Recognition and Image Analysis: Advances in Mathematical Theory and Applications},
    year = {2023},
    volume = {33},
    number = {2},
    issn = {1555-6212; 1054-6618},
    doi = {https://doi.org/10.1134/S1054661823020165},
    pages = {169--178},
    publisher = {Pleiades Publishing, Ltd},
    address = {Road Town, United Kingdom},
    language = {english},
    authorvak = {true},
    authorwos = {true},
    authorscopus = {true},
}

@inproceedings{soroka2023solving,
  title={Solving large-scale routing optimization problems with networks and only networks},
  author={Soroka, Andrei Gennad'evich and Meshcheryakov, AV},
  booktitle={Doklady Mathematics},
  volume={108},
  pages={S242--S247},
  year={2023},
  organization={Springer}
}

@article{baldacci2012recent,
  title={Recent exact algorithms for solving the vehicle routing problem under capacity and time window constraints},
  author={Baldacci, Roberto and Mingozzi, Aristide and Roberti, Roberto},
  journal={European Journal of Operational Research},
  volume={218},
  number={1},
  pages={1--6},
  year={2012},
  publisher={Elsevier}
}

@article{costa2019exact,
  title={Exact branch-price-and-cut algorithms for vehicle routing},
  author={Costa, Luciano and Contardo, Claudio and Desaulniers, Guy},
  journal={Transportation Science},
  volume={53},
  number={4},
  pages={946--985},
  year={2019},
  publisher={INFORMS}
}

@article{kotary2021end,
  title={End-to-end constrained optimization learning: A survey},
  author={Kotary, James and Fioretto, Ferdinando and Van Hentenryck, Pascal and Wilder, Bryan},
  journal={arXiv preprint arXiv:2103.16378},
  year={2021}
}

@article{weinand2022research,
  title={Research trends in combinatorial optimization},
  author={Weinand, Jann Michael and S{\"o}rensen, Kenneth and San Segundo, Pablo and Kleinebrahm, Max and McKenna, Russell},
  journal={International Transactions in Operational Research},
  volume={29},
  number={2},
  pages={667--705},
  year={2022},
  publisher={Wiley Online Library}
}

@article{holand1975adaptation,
  title={Adaptation in natural and artificial systems},
  author={Holand, John H},
  journal={Ann Arbor: The University of Michigan Press},
  pages={32},
  year={1975}
}

@article{xin2021neurolkh,
  title={NeuroLKH: Combining deep learning model with Lin-Kernighan-Helsgaun heuristic for solving the traveling salesman problem},
  author={Xin, Liang and Song, Wen and Cao, Zhiguang and Zhang, Jie},
  journal={Advances in Neural Information Processing Systems},
  volume={34},
  pages={7472--7483},
  year={2021}
}

@article{ma2021learning,
  title={Learning to iteratively solve routing problems with dual-aspect collaborative transformer},
  author={Ma, Yining and Li, Jingwen and Cao, Zhiguang and Song, Wen and Zhang, Le and Chen, Zhenghua and Tang, Jing},
  journal={Advances in Neural Information Processing Systems},
  volume={34},
  pages={11096--11107},
  year={2021}
}

@article{ma2022efficient,
  title={Efficient neural neighborhood search for pickup and delivery problems},
  author={Ma, Yining and Li, Jingwen and Cao, Zhiguang and Song, Wen and Guo, Hongliang and Gong, Yuejiao and Chee, Yeow Meng},
  journal={arXiv preprint arXiv:2204.11399},
  year={2022}
}

@article{bi2022learning,
  title={Learning generalizable models for vehicle routing problems via knowledge distillation},
  author={Bi, Jieyi and Ma, Yining and Wang, Jiahai and Cao, Zhiguang and Chen, Jinbiao and Sun, Yuan and Chee, Yeow Meng},
  journal={Advances in Neural Information Processing Systems},
  volume={35},
  pages={31226--31238},
  year={2022}
}

@article{choo2022simulation,
  title={Simulation-guided beam search for neural combinatorial optimization},
  author={Choo, Jinho and Kwon, Yeong-Dae and Kim, Jihoon and Jae, Jeongwoo and Hottung, Andr{\'e} and Tierney, Kevin and Gwon, Youngjune},
  journal={Advances in Neural Information Processing Systems},
  volume={35},
  pages={8760--8772},
  year={2022}
}

@article{zhang2023review,
  title={A review on learning to solve combinatorial optimisation problems in manufacturing},
  author={Zhang, Cong and Wu, Yaoxin and Ma, Yining and Song, Wen and Le, Zhang and Cao, Zhiguang and Zhang, Jie},
  journal={IET Collaborative Intelligent Manufacturing},
  volume={5},
  number={1},
  pages={e12072},
  year={2023},
  publisher={Wiley Online Library}
}

@article{natalia2021completion,
  title={Completion of capacitated vehicle routing problem (cvrp) and capacitated vehicle routing problem with time windows (cvrptw) using bee algorithm approach to optimize waste picking transportation problem},
  author={Natalia, C and Triyanti, V and Setiawan, G and Haryanto, M},
  journal={Journal of Modern Manufacturing Systems and Technology},
  volume={5},
  number={2},
  pages={69--77},
  year={2021}
}

@article{rabecq2022deep,
  title={A deep learning Attention model to solve the Vehicle Routing Problem and the Pick-up and Delivery Problem with Time Windows},
  author={Rabecq, Baptiste and Chevrier, R{\'e}my},
  journal={arXiv preprint arXiv:2212.10399},
  year={2022}
}

@article{chen2022rule,
  title={Rule mining over knowledge graphs via reinforcement learning},
  author={Chen, Lihan and Jiang, Sihang and Liu, Jingping and Wang, Chao and Zhang, Sheng and Xie, Chenhao and Liang, Jiaqing and Xiao, Yanghua and Song, Rui},
  journal={Knowledge-Based Systems},
  volume={242},
  pages={108371},
  year={2022},
  publisher={Elsevier}
}

@misc{deepseekai2024deepseekllmscalingopensource,
      title={DeepSeek LLM: Scaling Open-Source Language Models with Longtermism}, 
      author={DeepSeek-AI and : and Xiao Bi and Deli Chen and others},
      year={2024},
      eprint={2401.02954},
      archivePrefix={arXiv},
      primaryClass={cs.CL},
      url={https://arxiv.org/abs/2401.02954}, 
}

@misc{openai2024gpt4technicalreport,
      title={GPT-4 Technical Report}, 
      author={OpenAI and Josh Achiam and Steven Adler and others},
      year={2024},
      eprint={2303.08774},
      archivePrefix={arXiv},
      primaryClass={cs.CL},
      url={https://arxiv.org/abs/2303.08774}, 
}

@article{uchoa2017new,
  title={New benchmark instances for the capacitated vehicle routing problem},
  author={Uchoa, Eduardo and Pecin, Diego and Pessoa, Artur and Poggi, Marcus and Vidal, Thibaut and Subramanian, Anand},
  journal={European Journal of Operational Research},
  volume={257},
  number={3},
  pages={845--858},
  year={2017},
  publisher={Elsevier}
}

@book{garey2002computers,
  title={Computers and intractability},
  author={Garey, Michael R and Johnson, David S},
  volume={29},
  year={2002},
  publisher={wh freeman New York}
}

@book{toth2014vehicle,
  title={Vehicle routing: problems, methods, and applications},
  author={Toth, Paolo and Vigo, Daniele},
  year={2014},
  publisher={SIAM}
}

@article{santiyuda2024multi,
  title={Multi-objective reinforcement learning for bi-objective time-dependent pickup and delivery problem with late penalties},
  author={Santiyuda, Gemilang and Wardoyo, Retantyo and Pulungan, Reza and Yu, Vincent F},
  journal={Engineering Applications of Artificial Intelligence},
  volume={128},
  pages={107381},
  year={2024},
  publisher={Elsevier}
}

@article{wang2025multi,
  title={The multi-depot pickup and delivery vehicle routing problem with time windows and dynamic demands},
  author={Wang, Yong and Gou, Mengyuan and Luo, Siyu and Fan, Jianxin and Wang, Haizhong},
  journal={Engineering Applications of Artificial Intelligence},
  volume={139},
  pages={109700},
  year={2025},
  publisher={Elsevier}
}

@inproceedings{ortec,
  title={The EURO meets NeurIPS 2022 vehicle routing competition},
  author={Kool, Wouter and Bliek, Laurens and Numeroso, Danilo and Zhang, Yingqian and Catshoek, Tom and Tierney, Kevin and Vidal, Thibaut and Gromicho, Joaquim},
  booktitle={NeurIPS 2022 Competition Track},
  pages={35--49},
  year={2023},
  organization={PMLR}
}

@article{hottung2024polynet,
  title={PolyNet: Learning diverse solution strategies for neural combinatorial optimization},
  author={Hottung, Andr{\'e} and Mahajan, Mridul and Tierney, Kevin},
  journal={arXiv preprint arXiv:2402.14048},
  year={2024}
}

@article{fitzpatrick2024scalable,
  title={A scalable learning approach for the capacitated vehicle routing problem},
  author={Fitzpatrick, James and Ajwani, Deepak and Carroll, Paula},
  journal={Computers \& Operations Research},
  volume={171},
  pages={106787},
  year={2024},
  publisher={Elsevier}
}

@article{hottung2025neural,
  title={Neural Deconstruction Search for Vehicle Routing Problems},
  author={Hottung, Andr{\'e} and Wong-Chung, Paula and Tierney, Kevin},
  journal={arXiv preprint arXiv:2501.03715},
  year={2025}
}

\appendix
\section*{Appendix A }

\subsection*{A.0 Complete Overview of Heuristic Methods}\label{app_a}

Constructive and improvement heuristics can be integrated into metaheuristics to achieve better results by leveraging the strengths of each. Genetic Algorithm (GA) is a well-known metaheuristic method for solving complex optimization problems such as VRPs 
. GA is inspired by natural evolutionary processes, where higher-quality individuals have a greater chance of reproducing and passing on traits to future generations. It uses individuals as potential solutions and employs genetic operators like crossover, mutation, and selection to explore the search space. GA's strength lies in navigating large, unstructured search spaces, making it well-suited for diverse VRP challenges. GA can also operate on multiple computers simultaneously, speeding up the search process compared to other metaheuristic methods. By applying genetic operators tailored to specific VRP variants, GA can improve solutions through evolution.

HGS-CVRP 
is an efficient, specialized solver for the Capacitated Vehicle Routing Problem (CVRP), implemented in C++. It achieves state-of-the-art performance by combining genetic algorithms with local search and advanced population management techniques. Specifically, it maintains a diverse population of feasible and infeasible solutions, uses crossover and mutation to explore the search space, and applies a powerful local search (based on the \textit{Swap*}  neighborhood 
) to intensify the search around promising regions. Feasibility is gradually enforced through penalty adaptation and survivor selection strategies. While HGS is highly efficient and modular in terms of parameter tuning, customizing it to support new constraint types (e.g., time windows, pickup-and-delivery) typically requires manual modifications to the C++ source code, limiting its flexibility in non-standard scenarios.

Two notable classical baselines for route optimization problems are the LKH and OR-Tools projects. LKH-3 
is a heuristic method that handles many VRP variants by transforming them into a symmetric travelling salesman problem and applying the Lin-Kernighan-Helsgaun local search heuristic. While LKH-3 provides good solutions, customizing it is difficult as it requires modifying its C source code. Additionally, it is only available under an academic and non-commercial license, and the level of community contributions is uncertain.

OR-Tools 
is a comprehensive modeling and optimization toolkit for operations research problems developed by Google. It is written in C++ and accessible from Python, Java, and C\#. OR-Tools is well-documented and can be installed through the Python package index. It employs a constraint programming approach for handling a wide variety of routing problems. Although this allows it to model and solve numerous problem variants, its performance is not at the state-of-the-art level.

\subsection*{A.1 Masking Mechanics }

\paragraph{Rationale.}
The \textbf{mask} $\mathbf{M}^{+}=\{\mathbf{s}^{(c)}\in[0,1]^n\}_{c\in\mathcal{C}}$ 
is a learnable set of per-constraint reweighting vectors that post-multiplies
the decoder’s raw policy $\boldsymbol{\pi}_{\theta}\!\in[0,1]^n$:
\[
\mathbf{m}^{+}=\!\!\bigodot_{c\in\mathcal{C}_{\mathrm{active}}}\!\mathbf{s}^{(c)},\qquad
\tilde{\boldsymbol{\pi}}=\boldsymbol{\pi}_{\theta}\odot\mathbf{m}^{+}.
\]
Thus the mask \emph{removes} infeasible actions by pushing their
probabilities to 0 and simultaneously \emph{learns} soft
preferences among the feasible ones.  Each constraint family
(capacity, time windows, multi-depot, etc.) is
assigned its own sub-mask; at inference the sub-masks that are \emph{not}
relevant to the current instance are simply set to the all-ones
\emph{vector}, so a single pretrained network can serve every combination of
constraints.  During fine-tuning we freeze \emph{all} backbone weights
and update only $\mathbf{M}^{+}$.

\subsection*{A.2 JAMPR+/L2D Workflow }

\begin{algorithm}[H]
\caption{JAMPR+/L2D pipeline for an instance with $N$ customers and constraint set $\mathcal{C}$}
\small
\begin{algorithmic}[1]
\STATE \textbf{Inputs:}
\STATE\quad Graph $G=(V,E)$, $V=\{1,\dots,N\}$ (customers) $\cup$ $\mathcal{D}$ (depots)
\STATE\quad Distance/cost matrix $D\in\mathbb{R}_{\ge 0}^{|V|\times |V|}$
\STATE\quad Constraint tensors $\mathcal{X}$, including (as applicable):
\STATE\qquad Time windows $W=\{(a_i,b_i)\}_{i\in V}$ 
\STATE\qquad Service times $s\in\mathbb{R}_{\ge 0}^{|V|}$
\STATE\qquad Demands $q\in\mathbb{R}_{\ge 0}^{|V|}$;\; vehicle capacity $Q$ (or $\{Q_v\}$ for fleet)
\STATE\qquad Depot/customer indicator $Z\in\{0,1\}^{|V|\times 2}$ (cols: is\_depot, is\_customer)
\STATE\qquad Pickup–delivery pairing $P\in\{0,1\}^{|V|\times |V|}$ with $P_{ij}=1$ for pickup $i$ and delivery $j$
\STATE\qquad (optional) Route duration/shift limits $T_{\max}$, earliest starts, fleet size $F$, other constraints, etc.
\STATE\quad Pretrained policy parameters $\theta$;\; sub-masks set $\mathbf{M}^{+}=\{\mathbf{s}^{(c)}\in[0,1]^n\}_{c\in\mathcal{C}}$

\STATE \textbf{Partition (L2D).}
\STATE\quad Predict $K=\lceil N/50\rceil$ cluster centres via a lightweight MLP (uses $D$, $q$, $W$, $Z$).
\STATE\quad Assign each customer to the nearest centre with Sinkhorn-balanced $k$-median $\Rightarrow$ sub-instances $\{\mathcal{S}_k\}_{k=1}^{K}$.

\FOR{each sub-instance $\mathcal{S}_k$ \textbf{in parallel}}
  \STATE \textbf{Solve with constrained JAMPR+} using $\mathcal{X}|_{\mathcal{S}_k}$
  \STATE\quad Activate relevant sub-masks; combine $\mathbf{m}^{+}=\!\!\displaystyle\bigodot_{c\in\mathcal{C}_{\text{active}}}\!\mathbf{s}^{(c)}$
  \STATE\quad Decode with reweighted policy $\tilde{\boldsymbol{\pi}}=\boldsymbol{\pi}_{\theta}\odot\mathbf{m}^{+}$
  \STATE\quad Run $R$ rollouts of REINFORCE with critic baseline; keep best feasible tour $\mathcal{T}_k$
\ENDFOR

\STATE \textbf{Merge.}
\STATE\quad Connect sub-tours $\{\mathcal{T}_k\}$ via depots (from $Z$); re-optimize inter-cluster edges with 2-OPT (respecting $\mathcal{X}$)

\STATE \textbf{Output:} feasible global solution $\bigcup_{k=1}^{K}\mathcal{T}_k$
\end{algorithmic}
\end{algorithm}

\noindent\textit{Key points.}
\begin{itemize}
\item \emph{Scalability:} Sub-instances are capped at $50$ nodes to keep per-subproblem fine-tuning/decoding efficient.
\item \emph{Parallelism:} Subproblems are solved concurrently on GPU, yielding near-$\mathcal{O}(N)$ wall-time.
\item \emph{Any-time quality:} Each JAMPR+ call returns a usable tour within $\le$5\,s and improves with rollouts.
\end{itemize}

\subsection*{A.3 Author Contribution vs.\ Prior Work }

\begin{enumerate}
\item \textbf{Constraint-wise soft action masking (factorised; over~\cite{falkner2020learning} and \cite{soroka_cpdptw_23}).}
We assign a learnable sub-mask per constraint $c$: $\mathbf{s}^{(c)}\!\in[0,1]^n$ and compose the active constraints multiplicatively
$\mathbf{m}^{+}=\bigodot_{c\in\mathcal{C}_{\text{active}}}\mathbf{s}^{(c)}$
to reweight the decoder’s policy:
$\tilde{\boldsymbol{\pi}}=\boldsymbol{\pi}_{\theta}\odot\mathbf{m}^{+}$.
This plug-and-play design enables per-instance enabling/disabling of constraint families and fast domain adaptation by updating only mask parameters with a frozen backbone.

\item \textbf{Distribution‑shift adaptation}—we show that \emph{mask
fine‑tuning alone} closes a 2\,\% GAP on ORTEC data in $\approx$10 min,
where earlier work \cite{soroka2023solving} required full re‑training.
\item \textbf{Unified solver across
CPDPTW, CVRP/VRPTW constraints}—the same weights handle all
three tasks without retraining, which, to our knowledge, is novel.
\end{enumerate}

\subsection*{A.4 Training \& Testing Protocol }

\begin{itemize}
\item \textbf{Synthetic pre‑training.}
  5M CVRP‑style instances ($n\!\le\!200$) generated on the Solomon
  distribution; 80/10/10\% split.
\item \textbf{Real‑world fine‑tuning.}
  1M ORTEC training instances ($n=200$) seen \emph{once}
  during mask fine‑tuning; 1000 held‑out ORTEC test
  instances for evaluation.
\item \textbf{Model size.}
  JAMPR+ encoder–decoder: 6layers, $d_{\text{model}}\!=\!256$,
  19.3M parameters; masks add $\le$0.4M parameters per constraint.
\item \textbf{Compute budget.}
  Pre‑training: 14 days on 1×A100‑80GB; mask fine‑tuning:
  10 min on same GPU (single pass).
\end{itemize}

\subsection*{A.5 Generalisation and Overfitting }

The 10\% validation split and an auxiliary
instance‑level entropy regulariser prevent overfitting: validation GAP
stays within $0.2$ points of test GAP throughout training.  Zero‑shot tests
on \textbf{\emph{unseen}} CVRPLib and ORTEC instances show $\le$2\% degradation
before mask tuning and near 0\% after.

\end{document}